\documentclass[11pt,a4paper]{article}
\usepackage[hyperref]{emnlp2020}
\usepackage{times}
\usepackage{latexsym}

\usepackage{microtype}

\aclfinalcopy 
\def\aclpaperid{590} 

\usepackage{amsfonts}
\usepackage{amsmath}
\usepackage{amsthm}
\usepackage{xspace}
\usepackage{color}
\usepackage{array}
\usepackage{multirow}
\usepackage{longtable}
\usepackage{floatrow}
\usepackage[normalem]{ulem}
\usepackage{booktabs}
\usepackage{graphicx}
\usepackage{bm}
\usepackage{bbm}
\usepackage{enumitem}
\usepackage{floatrow}
\newfloatcommand{capbtabbox}{table}[][\FBwidth]

\newcommand*{\eg}{e.g.\@\xspace}
\newcommand*{\ie}{i.e.\@\xspace}

\title{A Benchmark for Structured Procedural Knowledge Extraction \\ from Cooking Videos}

\author{
Frank F. Xu$^1$\thanks{~~Work done during the first author's internship at Microsoft Research, Asia. Data and code: \url{https://github.com/frankxu2004/cooking-procedural-extraction}.}, Lei Ji$^2$, Botian Shi$^3$, Junyi Du$^4$, Graham Neubig$^1$, Yonatan Bisk$^1$, Nan Duan$^2$ \\
$^1$Carnegie Mellon University $^2$Microsoft Research Asia  \\
$^3$Beijing Institute of Technology $^4$University of Southern California \\
\texttt{\{fangzhex,gneubig,ybisk\}@cs.cmu.edu}, \texttt{\{leiji,nanduan\}@microsoft.com}
}

\date{}

\begin{document}

\maketitle

\begin{abstract}
Watching instructional videos are often used to learn about procedures. 
Video captioning is one way of automatically collecting such knowledge.
However, it provides only an indirect, overall evaluation of multimodal models with no finer-grained quantitative measure of what they have learned.  
We propose instead, a benchmark of \emph{structured} procedural knowledge extracted from cooking videos.  
This work is complementary to existing tasks, but requires models to produce interpretable structured knowledge in the form of verb-argument tuples.
Our manually annotated open-vocabulary resource includes 356 instructional cooking videos and 15,523 video clip/sentence-level annotations.  
Our analysis shows that the proposed task is challenging and standard modeling approaches like unsupervised segmentation, semantic role labeling, and visual action detection perform poorly when forced to predict every action of a procedure in structured form.


\end{abstract}

\section{Introduction}

Instructional videos are  a convenient way to  learn a new skill.
Although learning from video seems natural to humans, it requires identifying and understanding procedures and grounding them to the real world.
In this paper, we propose a new task and dataset for extracting procedural knowledge into a fine-grained \emph{structured} representation from \emph{multimodal} information contained in a \emph{large-scale} archive of \emph{open-vocabulary} narrative videos with \emph{noisy transcripts}.
While there is a significant amount of related work (summarized in \S\ref{sec:dataset} \& \ref{sec:related}), to our knowledge there is no dataset similar in scope, with previous attempts focusing only on a single modality (e.g.,~text only \citep{Kiddon2015MiseEP} or video only \citep{zhukov2019cross,alayrac2016unsupervised}), using closed-domain taxonomies \citep{tang2019coin}, or lacking structure in the procedural representation \citep{zhou2018weakly}.

\begin{figure}[t]
  \centering
  \includegraphics[width=0.9\linewidth]{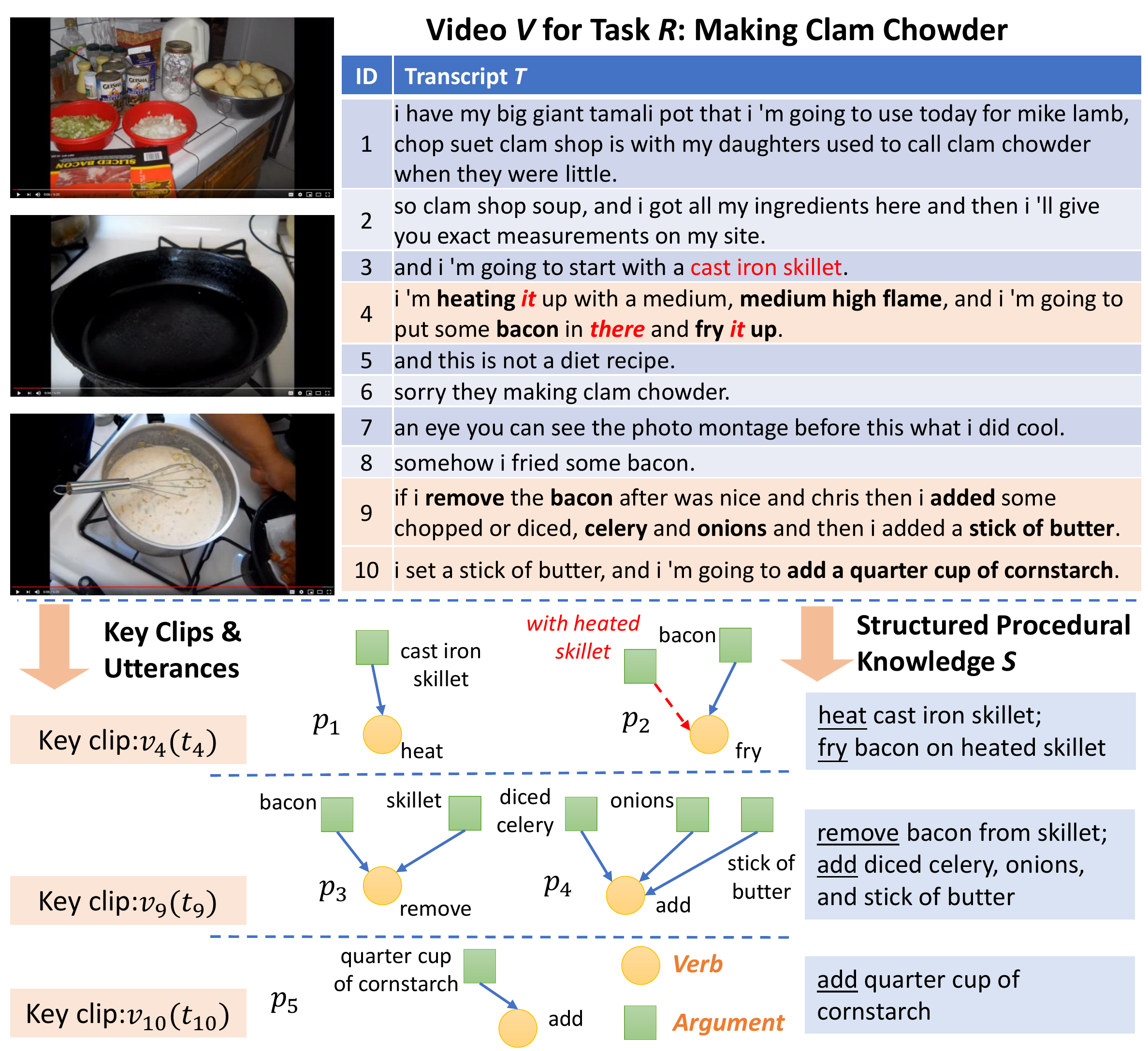}
  \small
  \caption{An example of extracting procedures for task \emph{``Making Clam Chowder''}.
  }
  \label{fig:example}
\end{figure}

In our task, given a narrative video, say a cooking video on YouTube about \emph{making clam chowder} as shown in Figure~\ref{fig:example}, our goal is to extract a series of tuples representing the procedure, \eg (heat, cast iron skillet), (fry, bacon, with heated skillet), etc.
We created a manually annotated, large test dataset for evaluation of the task, including over 350 instructional cooking videos along with over 15,000 English sentences in the transcripts spanning over 89 recipe types.
This verb-argument structure using arbitrary textual phrases is motivated by open information extraction~\cite{schmitz2012open,fader2011identifying}, 
but focuses on procedures rather than entity-entity relations.

This task is challenging with respect to both video and language understanding.
For video, it requires understanding of video contents, with a special focus on actions and procedures.
For language, it requires understanding of oral narratives, including understanding of predicate-argument structure and coreference.
In many cases it is necessary for both modalities to work together, such as when resolving null arguments necessitates the use of objects or actions detected from video contents in addition to transcripts. 
For example, the cooking video host may say ``just a pinch of salt in'', while adding some salt into a boiling pot of soup, in which case inferring the action  ``add'' and its argument ``pot'' requires visual understanding. 
Along with the novel task and dataset, we propose several baseline approaches that extract structure in a pipelined fashion.
These methods first identify key clips/sentences using video and transcript information with unsupervised and supervised multimodal methods, then extract procedure tuples from the utterances and/or video of these key clips.
On the utterances side, we utilize an existing state-of-the-art semantic role labeling model~\cite{shi2019simple}, with the intuition that semantic role labeling captures the verb-argument structures of a sentence, which would be directly related to procedures and actions.
On the video side, similarly, we utilize existing state-of-the-art video action/object recognition model trained in kitchen settings to further augment utterance-only extraction results.
The results are far from perfect, demonstrating that the proposed task is challenging and that structuring procedures requires more than just state-of-the-art semantic parsing or video action recognition.



\section{Problem Definition}
\label{sec:def}
We show a concrete example of our procedural knowledge extraction task in Figure~\ref{fig:example}. 
Our ultimate goal is to automatically map \textit{unstructured} instructional video (clip and utterances) to \textit{structured} procedures, defining what actions should be performed on
which objects, with what arguments and in what order. 
We define the input to such an extraction system:
\vspace{-0.2cm}
\begin{itemize}[leftmargin=5mm]
\setlength\itemsep{-1mm}
    \item Task $R$, \eg ``Create Chicken Parmesan'' and instructional video $V_R$ describing the procedure to achieve task $R$, \eg a video titled ``Chicken Parmesan - Let's Cook with ModernMom''.\footnote{\url{https://www.youtube.com/watch?v=nWGpCmDlNU4}}
    \item A sequence of $n$ sentences $T_R = \{t_0, t_1, ..., t_n\}$ representing video $V_R$'s corresponding transcript. According to the time stamps of the transcript sentences, the video is also segmented into $n$ clips $V_R = \{v_0, v_1, ..., v_n\}$ accordingly to align with the sentences in the transcript $T_R$.
\end{itemize}

\noindent
The output will be: 
\begin{itemize}[leftmargin=4mm]
\setlength\itemsep{-2mm}
    \item A sequence of $m$ procedure tuples $S_R = \{s_0, s_1, ..., s_m\}$ describing the key steps to achieve task $R$ according to instructional video $V_R$.
    \item An identified list of \emph{key} video clips and corresponding sentences $V'_R \subseteq V_R$, to which procedures in $S_R$ are grounded.
\end{itemize}
Each procedural tuple
$s_j  = (\text{verb}, \text{arg}_1, ..., \text{arg}_k) \in S_R$
consists of a verb phrase and its arguments. Only the ``verb'' field is required, and thus the tuple size ranges from 1 to $k+1$. 
All fields can be either a word or a phrase.

Not every clip/sentence describes procedures, as most videos include an intro, an outro, non-procedural narration, or off-topic chit-chat.
Key clips $V'_R$ are clips associated with one or more procedures in $P_R$, with some clips/sentences associated with multiple procedure tuples. 
Conversely, each procedure tuple will be associated with only a single clip/sentence.

\section{Dataset \& Analysis}

\label{sec:dataset}
\begin{table}[t]
\centering
\resizebox{\textwidth}{!}{%
\small
\begin{tabular}{@{}lc@{\hspace{0.5em}}c@{\hspace{0.5em}}c@{\hspace{0.5em}}c@{\hspace{0.5em}}c@{\hspace{0.5em}}c@{\hspace{0.5em}}c@{\hspace{0.5em}}c@{}}
\toprule
                   & Ours       & AR         & YC2        & CT  & COIN & How2  & HAKE & TACOS    \\ 
                   \midrule
General domain?    &            &            &            & \checkmark & \checkmark & \checkmark & \checkmark & \\
Multimodal input?  & \checkmark &            &            &            &        & \checkmark & \checkmark & \checkmark   \\
Use transcript?    & \checkmark &            &            &            &    &  \checkmark &    &    \\
Use noisy text?    & \checkmark &            &            &            &    & \checkmark  &    &    \\
Open extraction?   & \checkmark &            & \checkmark &            &      &   &    &      \\
Structured format? & \checkmark & \checkmark &            & \checkmark &        &   & \checkmark & \checkmark  \\
\bottomrule
\end{tabular}%
}
\caption{Comparison to current datasets.}
\label{tab:datasetfeature}
\end{table}

While others have created related datasets, they fall short on key dimensions which we remedy in our work.  Specifically, In Table~\ref{tab:datasetfeature} we compare to AllRecipes~\cite{Kiddon2015MiseEP} (AR), YouCook2~\cite{zhou2018towards} (YC2),  CrossTask~\cite{zhukov2019cross} (CT), COIN~\cite{tang2019coin}, How2~\cite{sanabria2018how2}, HAKE~\cite{li2019hake} and TACOS~\cite{regneri2013grounding}.  
Additional details about datasets are included in the Appendix \ref{app:existing_datasets}.%
\footnote{A common dataset we do not include here is HowTo100M \cite{miech19howto100m} as it does not contain any annotations.} 
In summary, none have both \emph{structured} and \emph{open} extraction annotations for the procedural knowledge extraction task,
since most focus on either video summarization/captioning or action localization/classification.

\begin{figure}[t]
  \centering
  \includegraphics[width=0.80\linewidth]{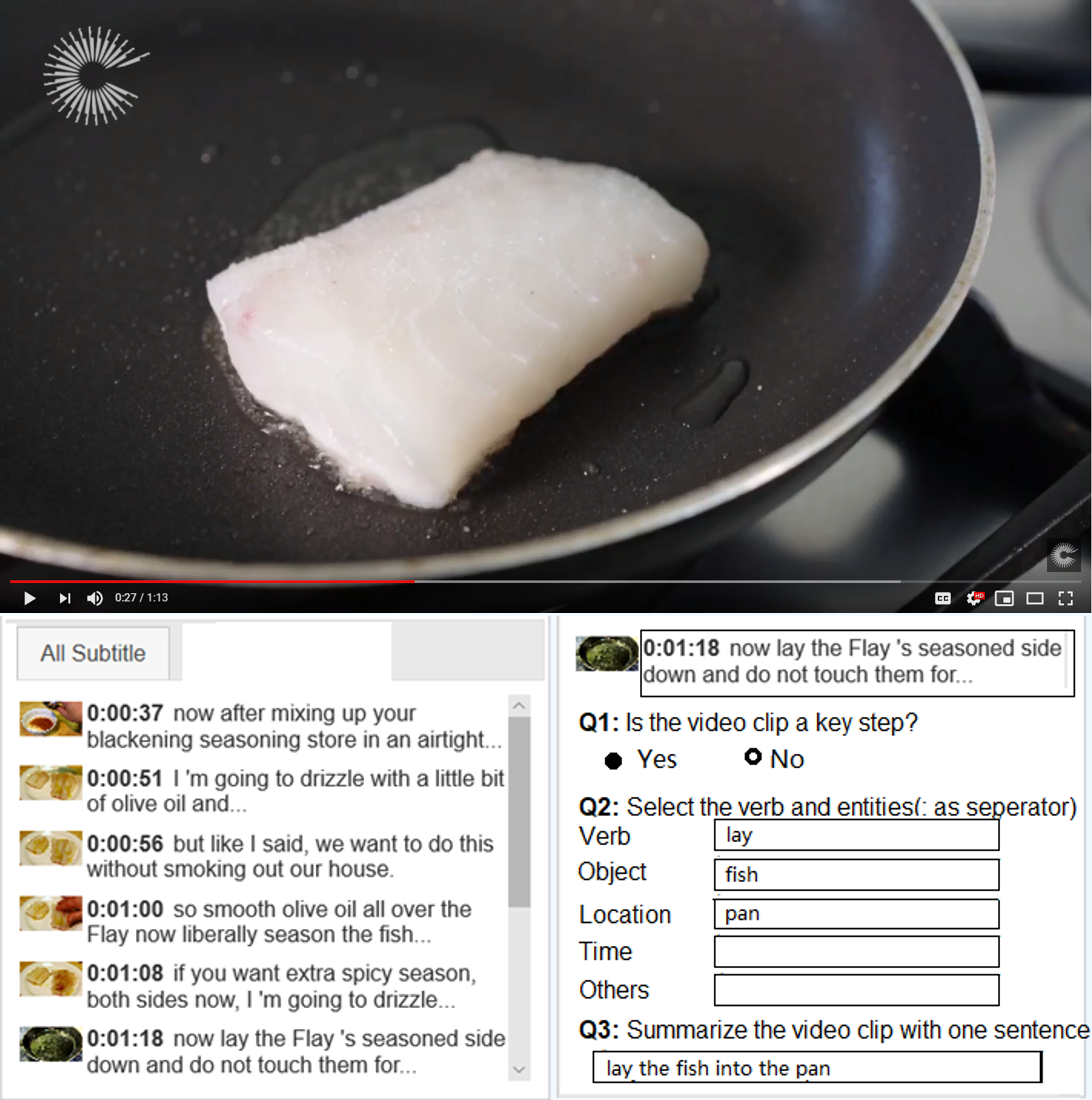}
  \caption{Annotation interface.}
  \label{fig:annotation_tool}
\end{figure}

\subsection{Dataset Creation}
To address the limitations of existing datasets, we created our own evaluation dataset by annotating structured procedure knowledge given the video and transcript. 
Native English-speakers annotated four videos per recipe type (\eg clam chowder, pizza margherita, etc.) in the YouCook2 dataset into the structured form presented in \S\ref{sec:def} (totaling 356 videos). 
Annotators selected key clips as important steps and extracted corresponding fields to fill in verbs and arguments. 
Filling in the fields with the original tokens was preferred but not required (e.g., in cases of coreference and ellipsis).
The result is a series of video clips labeled with procedural structured knowledge as a sequence of steps $s_j$ and series of short sentences describing the procedure.

\begin{table}[t]
\centering
\caption{Statistics of annotated verbs and arguments in procedures.}
\label{tab:annotationstat}
\small
\begin{tabular}{lcc}
\hline
                            & Verbs & Arguments \\ \hline
Total \#                    & 4004  & 6070      \\ \hline
Average \# per key clip     & 1.12  & 1.70      \\ \hline
Average \#words             & 1.07  & 1.43      \\ \hline
\% directly from transcript & 69.8  & 75.0      \\ \hline
\% coreference (pronouns) & N/A   & 14.4      \\ \hline
\% ellipsis                 & 30.2  & 10.6      \\ \hline
\end{tabular}
\end{table}

Figure~\ref{fig:annotation_tool} shows the user interface of annotation tool. The  process is divided into 3 questions per clip:
\textbf{Q1:} Determine if the video clip is a key step if:
(1) the clip or transcript contains at least one action; 
(2) the action is required for accomplishing the task (\ie not a self introduction); 
and (3) for if a clip duplicates a previous key clip, choose the one with clearer visual and textual signals (\eg without coreference, etc.).
\textbf{Q2:} For each key video clip, annotate the key procedural tuples. 
We have annotators indicate which actions are both seen and mentioned by the instructor in the video.
The actions should correspond to a verb and its arguments from the original transcript except in the case of ellipsis or coreference where they have to refer to earlier phrases based on the visual scene.
\textbf{Q3:} Construct a short fluent sentence from the annotated tuples 
for the given video clip. 

We have two expert annotators and a professional labeling supervisor for quality control and deciding the final annotations. 
To improve the data quality, the supervisor reviewed all labeling results, and applied several heuristic rules to find anomalous records for further correction. 
The heuristic is to check the annotated verb/arguments that are not found in corresponding transcript text. 
Among these anomalies, the supervisor checks the conflicts between the two annotators. 
25\% of all annotations were modified as a result.
On average annotators completed task Q1 at 240 sentences (clips) per hour and task Q2 and Q3 combined at 40 sentences per hour.
For Q1, we observe an inter-annotator agreement with Cohen's Kappa of 0.83.\footnote{We use the Jaccard ratio between the annotated tokens of two annotators for Q2's agreement. Verb annotations have a higher agreement at 0.77 than that of arguments at 0.72.}
Examples are shown in Table~\ref{tab:annotationexamples}. 

\begin{table*}[t]
\caption{
Annotations of structured procedures and summaries. \emph{Coreference} and \emph{ellipsis} are marked with \emph{italics} and are resolved into referred phrases also linked back in the annotations. See Appendix (Table~\ref{tab:casestudy}) for more examples.}
\label{tab:annotationexamples}
\resizebox{\textwidth}{!}{%
\small
\begin{tabular}{p{6cm}|p{4cm}|p{1cm}|p{2.2cm}|p{2cm}}
Transcript sentence                                                                                         & Procedure summary                 & Verb    & \multicolumn{2}{p{4cm}}{Arguments}  \\ \hline
so we've \textbf{placed the dough directly into the caputo flour} that we import from italy.  & place dough in caputo flour      & place   & dough           & caputo flour  \\ \hline
\multirow{2}{\hsize}{we just \textbf{give \emph{(ellipsis)} a squish with our palm and make \emph{it} flat in the center}.}             & squish dough with palm           & squish  & dough           & with palm     \\ \cline{2-5} 
                                                                                                 & flatten center of dough          & flatten & center of dough &               \\ \hline
so will have to \textbf{rotate \emph{it} every thirty to forty five seconds} ...                                 & rotate pizza every 30-45 seconds & rotate  & pizza           & every 30-45 seconds
\end{tabular}%
}
\end{table*}

\begin{figure}[t]
  \centering
  \includegraphics[width=0.8\linewidth]{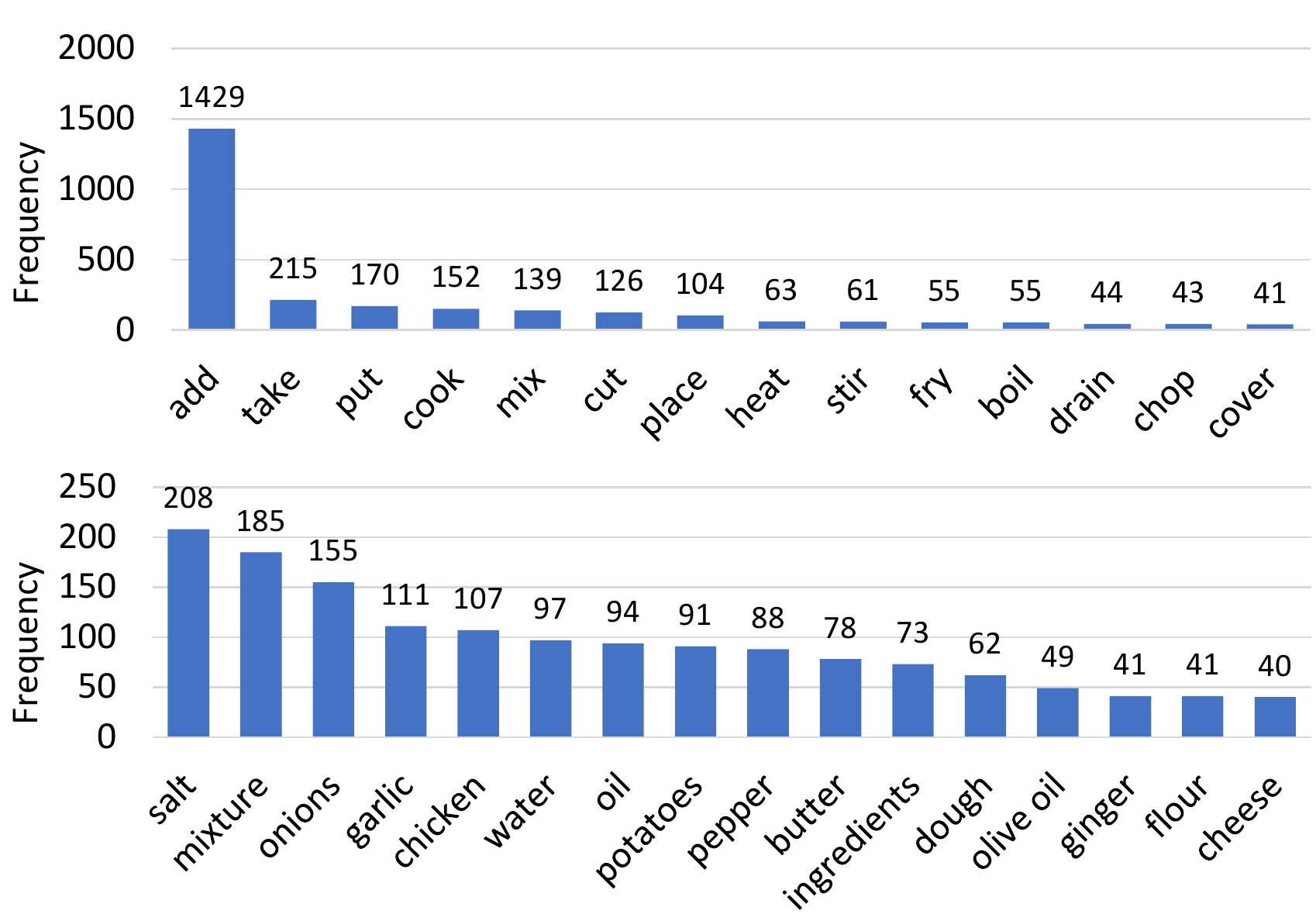}
  \caption{Most frequent verbs (upper) and arguments (lower).}
  \label{fig:freq_list}
\end{figure}

\subsection{Dataset Analysis}
\label{sec:analysis}

Overall, the dataset contains 356 videos with 15,523 video clips/sentences, among which 3,569 clips are labeled as key steps. 
Sentences average 16.3 tokens, and the language style is oral English.
For structured procedural annotations, there are 347 unique verbs and 1,237 unique objects in all.
Statistics are shown in Table~\ref{tab:annotationstat}.
Figure~\ref{fig:freq_list} lists the most commonly appearing verbs and entities. 
The action \emph{add} is most frequently performed, and the entities \emph{salt} and \emph{onions} are the most popular ingredients.

In nearly 30\% of annotations, some verbs and arguments cannot be directly found in the transcript.
An example is ``(add) some salt into the pot'', and we refer to this variety of absence as \emph{ellipsis}.
Arguments not mentioned explicitly are mainly due to (1) pronoun references, \eg ``put it (fish) in the pan''; (2) ellipsis, where the arguments are absent from the oral language, \eg ``put the mixture inside'' where the argument ``oven'' is omitted.
The details can be found in Table~\ref{tab:annotationstat}.
The coreferences and ellipsis phenomena add difficulty to our task, and indicate the utility of using multimodal information from the video signal and contextual procedural knowledge for inference. 

\begin{figure}[t]
  \centering
  \includegraphics[width=0.5\textwidth]{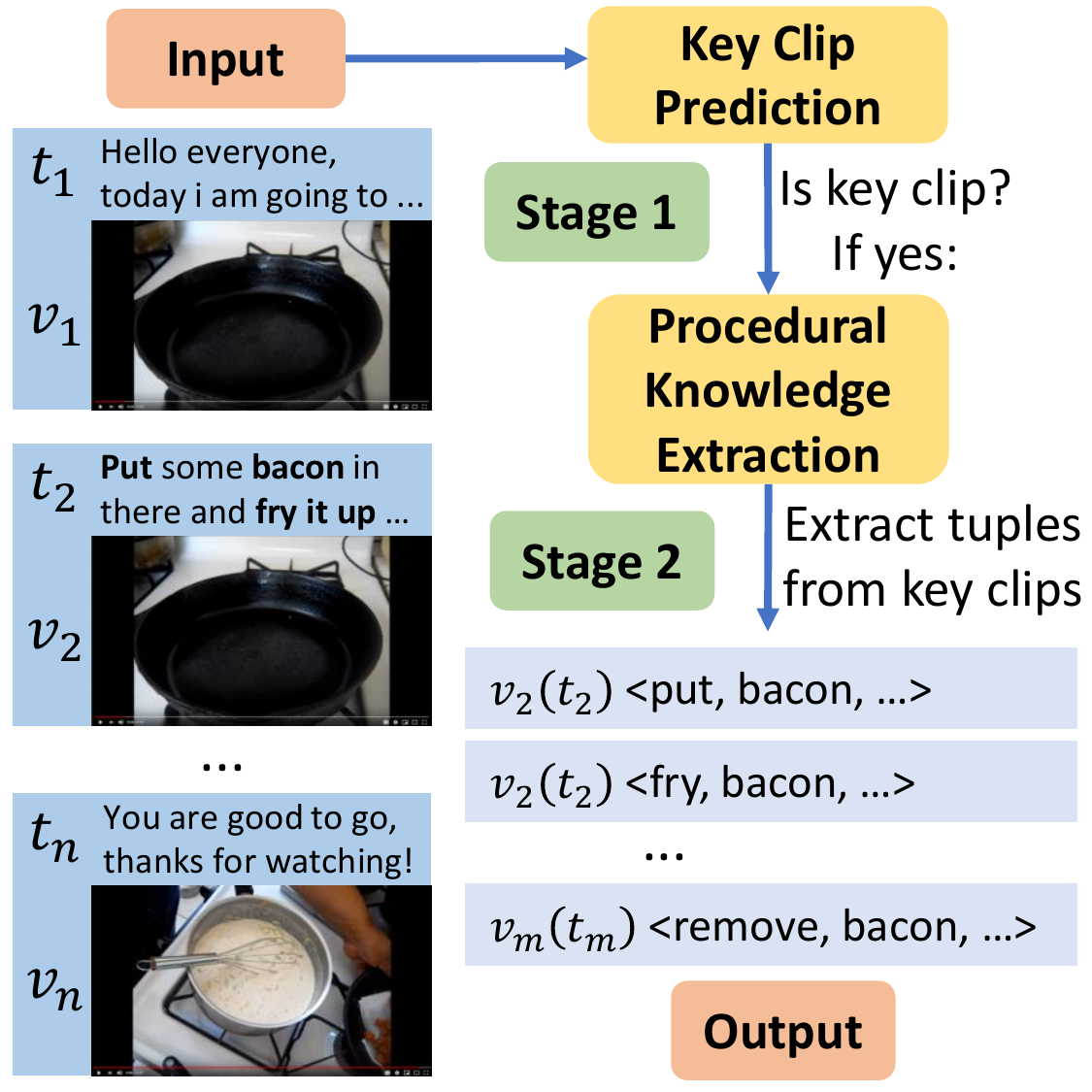}
  \caption{Extraction pipeline.}
  \label{fig:model_struct}
\end{figure}

\section{Extraction Stage 1: Key Clip Selection}
\label{sec:keyclipbaselines}
In this and the following section, we describe our two-step pipeline for procedural knowledge extraction (also in Figure~\ref{fig:model_struct}).
This section describes the first stage of determining which clips are ``key clips'' that contribute to the description of the procedure.
We describe several key clip selection models, which consume the transcript and/or the video within the clip and decide whether it is a key clip. 

\subsection{Parsing-Based Heuristic Baselines}
\label{sec:heuistic}
Given our unsupervised setting, we first examine two heuristic parsing-based methods that focus on the transcript only, one based on semantic role labeling (SRL) and the other based on an unsupervised segmentation model \citet{Kiddon2015MiseEP}.

Before introducing heuristic baselines, we note that having a lexicon of domain-specific actions will be useful, e.g.,~for filtering pretrained model outputs, or providing priors to the unsupervised model described later.
In our cooking domain, these actions can be expected to consist mostly of verbs related to cooking actions and procedures. 
Observing recipe datasets such as AllRecipes~\cite{Kiddon2015MiseEP} or WikiHow~\cite{miech19howto100m,zhukov2019cross}, we find that they usually use imperative and concise sentences for procedures and the first word is usually the action verb like \emph{``add''}, e.g.,~\emph{add some salt into the pot}. 
We thus construct a cooking lexicon by aggregating the frequently appearing verbs as the first word from AllRecipes, with frequency over a threshold of 5.
We further filter out words that have no verb synsets in WordNet~\cite{miller1995wordnet}. 
Finally we manually filter out noisy or too general verbs like ``go''.
Note that when applying to other domains, the lexicon can be built following a similar process of first finding a domain-specific corpus with simple and formal instructions, and 
then obtaining the lexicon by aggregation and filtering.

\noindent \textbf{Semantic role labeling baselines.} 
One intuitive trigger in the transcript for deciding whether the sentence is a key step should be the action words, \ie the verbs.
In order to identify these action words we use semantic role labeling~\cite{gildea2002automatic}, which analyzes natural language sentences to extract information about ``who did what to whom, when, where and how?''
The output is in the form of predicates and their respective arguments that acts as semantic roles, where the verb acts as the root (head) of the parse. 
We run a strong semantic role labeling model~\cite{shi2019simple}
included in the AllenNLP toolkit~\cite{gardner2018allennlp} on each sentence in the transcript. 
From the output we get a set of verbs for each of the sentences.%
\footnote{The SRL model is used in this stage only as a verb identifier, with other output information used in stage 2.}
Because not all verbs in all sentences represent actual key actions for the procedure, we additionally filter the verbs with the heuristically created cooking lexicon above, counting a clip as a key clip only if at least one of the SRL-detected verbs is included in the lexicon.

\noindent \textbf{Unsupervised recipe segmentation baseline~\cite{Kiddon2015MiseEP}.}
The second baseline is based on the outputs of the unsupervised recipe sentence segmentation model in~\citet{Kiddon2015MiseEP}. 
Briefly speaking, the model is a generative probabilistic model where verbs and arguments, together with their numbers, are modeled as latent variables.
It uses a bigram model for string selection. 
It is trained on the whole transcript corpus of YouCook2 videos iteratively for 15 epochs using a hard EM approach before the performance starts to converge.
The count of verbs in the lexicon created in \S\ref{sec:heuistic} is provided as a prior through initialization. 
We then do inference to parse the transcripts in our dataset using the trained model. 
Following the same heuristics as the SRL outputs, we treat sentences with non-empty parsed predicates after lexical filtering as key sentences, and those without as negatives.

\subsection{Neural Selection Baseline}
\label{sec:neuralsel}
Next, we implement a supervised neural network model that incorporates visual information, which we have posited before may be useful in the face of incomplete verbal utterances.
We extract the features of the sentence and each video frame using pretrained feature extractors respectively. 
Then we perform attention~\cite{bahdanau2014neural} over each frame feature, using the sentence as a query, in order to acquire the representation of the video clip.
Finally, we combine the visual and textual features to predict whether the input is a key clip. 
The model is trained on a \emph{general domain} instructional key clip selection dataset with \emph{no} overlap with ours, and our annotated dataset is used for evaluation \emph{only}.
Additional details about the model and training dataset are included in Appendix \ref{app:neural_selection_model}.

\section{Extraction Stage 2: Structured Knowledge Extraction}

With the identified key clips and corresponding transcript sentences, we proceed to the second stage that performs clip/sentence-level procedural knowledge extraction from key clips.
In this stage, the extraction is done from clips that are identified at first as “key clips”.

\subsection{Extraction From Utterances}
\label{sec:extract_utterances}
We first present two baselines to extract structured procedures using transcripts only, similarly to the key-clip identification methods described in \S\ref{sec:heuistic}. 

\noindent \textbf{Semantic role labeling.}
For the first baseline, we use the same pretrained SRL model introduced in \S\ref{sec:heuistic} to conduct inference on the sentences in key clips identified from stage 1. 
Because they consist of verb-argument structures, the outputs of the SRL model are well aligned with the task of extracting procedural tuples that identify actions and their arguments.
However, not all outputs from the SRL model are the structured procedural knowledge we aim to extract.
For example, in the sentence \emph{``you 're ready to add a variety of bell peppers''} from the transcript, the outputs from SRL model contains two parses with two predicates, \emph{``are''} and \emph{``add''}, where only the latter is actually part of the procedure.
To deal with this issue we first perform filtering similar to that used in stage 1, removing parses with predicates (verbs) outside of the domain-specific action lexicon we created in \S\ref{sec:heuistic}. 
Next, we filter out irrelevant arguments in the parse. 
For example, the parse from the SRL model for sentence ``I add a lot of pepper because I love it.'' after filtering out irrelevant verb ``love'' is ``[\texttt{ARG0}: I] [\texttt{V}: add] [\texttt{ARG1}: a lot of pepper] [\texttt{ARGM-CAU}: because I love it]'', some arguments such as \texttt{ARG0} and \texttt{ARGM-CAU} are clearly not contributing to the procedure.
We provide a complete list of the filtered argument types in Appendix~\ref{app:argument_filtering}.

\noindent \textbf{Unsupervised recipe segmentation~\cite{Kiddon2015MiseEP}.}
The second baseline is to use the same trained segmentation model as in \S\ref{sec:heuistic} to segment selected key transcript sentences into verbs and arguments.
We treat segmented predicates in the key sentence as procedural verbs, and segmented predicate arguments plus preposition arguments as procedural arguments.

\subsection{Extraction From Video}
\label{sec:videodetection}
We also examine a baseline that utilizes two forms of visual information in videos: actions and objects.
We predict both verbs and nouns of a given video clip via a state-of-the-art action detection model TSM~\cite{lin2019tsm},\footnote{\url{https://github.com/epic-kitchens/action-models}} trained on the EpicKitchen~\cite{Damen2018EPICKITCHENS} dataset.\footnote{\url{https://epic-kitchens.github.io/2019}}
For each video, we extract 5-sec video segments and feed into the action detection model. 
The outputs of the models are in a \emph{predefined} set of labels of verbs (actions) and nouns (objects).%
\footnote{Notably, this contrasts to our setting of attempting to recognize into an open label set, which upper-bounds the accuracy of any model with a limited label set.}
We directly combine the outputs from the model on each video segment, aggregate and temporally align them with key clips/sentences, forming the final output.

\subsection{Utterance and Video Fusion}
Finally, to take advantage of the fact that utterance and video provide complementary views, we perform multimodal fusion of the results of both of these model varieties.
We adopt a simple method of fusion by taking the union of the verbs/actions and arguments/objects respectively from the best performing utterance-only model and the visual detection model. 

\vspace{-0.2cm}
\section{Evaluation}
\vspace{-0.2cm}
We propose evaluation metrics and provide evaluation results on our annotated dataset for both of the two stages: key clip selection and structured procedural extraction.
Detailed reproducibility information about the experiments are in Appendix~\ref{app:repro}.
Besides quantitative evaluation and qualitative evaluations, 
we also analyze the key challenges of this task.

\begin{table}[t]
\centering
\caption{Key clip selection results.}
\label{tab:keysent}
\small
\begin{tabular}{c c c c c}
\hline
              & Acc   & P   & R   & F1    \\ \hline \hline
\multicolumn{5}{c}{Parsing-based Heuristics} \\ \hline
SRL w/o heur.              & 25.9 & 23.4 & \textbf{97.6} & 37.7 \\ \hline
SRL w/ heur.          & 61.2 & 35.2 & 81.4 & 49.1 \\ \hline
\citet{Kiddon2015MiseEP} & 67.3 & 33.5 & 42.7 & 37.6 \\ \hline
\multicolumn{5}{c}{Neural Model} \\ \hline
Visual Only      & 43.8 & 27.2 & 85.9 & 41.3 \\ \hline
Text Only        & 76.3 & 49.0 & 78.1 & 60.2 \\ \hline
V+T (Full Model) & \textbf{77.7} & \textbf{51.0} & 75.3 & \textbf{60.8} \\ \hline
\end{tabular}
\end{table}

\subsection{Extraction Stage 1: Key Clip Selection}
\label{sec:keyclipeval}

\begin{table*}[t]
\centering
\caption{\vspace{-5mm}Clip/sentence-level structured procedure extraction results for verbs and arguments.}
\label{tab:verbextract}
\setlength\itemsep{-5mm}
\resizebox{\textwidth}{!}{%
\begin{tabular}{ccccccccccccccccccc}
\hline
\multicolumn{1}{c|}{\multirow{3}{*}{Model}}   & \multicolumn{9}{c|}{Verbs}                                                                                                                                                                                   & \multicolumn{9}{c}{Arguments}                                                                                                                                                        \\ \cline{2-19} 
\multicolumn{1}{c|}{}                         & \multicolumn{3}{c|}{Exact Match}                                   & \multicolumn{3}{c|}{Fuzzy}                                   & \multicolumn{3}{c|}{Partial Fuzzy}                                 & \multicolumn{3}{c|}{Exact Match}                                & \multicolumn{3}{c}{Fuzzy}                                    & \multicolumn{3}{|c}{Partial Fuzzy}            \\ \cline{2-19} 
\multicolumn{1}{c|}{}                         & P             & R             & \multicolumn{1}{c|}{F1}            & P             & R             & \multicolumn{1}{c|}{F1}            & P             & R             & \multicolumn{1}{c|}{F1}            & P            & R            & \multicolumn{1}{c|}{F1}           & P             & R             & \multicolumn{1}{c|}{F1}            & P             & R             & F1            \\ \hline
\multicolumn{19}{c}{Using oracle key clips}                                                                                                                                                                                                                                                                                                                                                                                                         \\ \hline
\multicolumn{1}{c|}{\citet{Kiddon2015MiseEP}} & 12.0          & 10.9          & \multicolumn{1}{c|}{11.4}          & 18.8          & 17.2          & \multicolumn{1}{c|}{18.0}          & 20.2          & 18.4          & \multicolumn{1}{c|}{19.3}          & 0.4          & 0.9          & \multicolumn{1}{c|}{0.5}          & 10.4          & 19.3          & \multicolumn{1}{c|}{13.5}          & 16.4          & 30.2          & 21.3          \\ \hline
\multicolumn{1}{c|}{SRL w/o heur.}            & 19.4          & 54.7          & \multicolumn{1}{c|}{28.6}          & 25.3          & 70.1          & \multicolumn{1}{c|}{37.2}          & 26.6          & 73.8          & \multicolumn{1}{c|}{39.1}          & 1.3          & \textbf{5.4} & \multicolumn{1}{c|}{2.0}          & 14.1          & \textbf{53.6} & \multicolumn{1}{c|}{22.3}          & 22.0          & \textbf{81.8} & 34.6          \\ \hline
\multicolumn{1}{c|}{SRL w/ heur.}             & \textbf{38.7}          & 51.6          & \multicolumn{1}{c|}{\textbf{44.3}}          & \textbf{45.2}          & 60.3          & \multicolumn{1}{c|}{\textbf{51.7}}          & \textbf{46.9}          & 62.6          & \multicolumn{1}{c|}{\textbf{53.6}}          & \textbf{1.6}          & 3.3          & \multicolumn{1}{c|}{\textbf{2.2}}          & \textbf{21.2}          & 39.8          & \multicolumn{1}{c|}{\textbf{27.7}}          & \textbf{32.3}          & 59.5          & \textbf{41.9}          \\ \hline
\multicolumn{1}{c|}{Visual}                   & 4.1           & 6.7           & \multicolumn{1}{c|}{5.1}           & 17.9          & 27.8          & \multicolumn{1}{c|}{21.7}          & 19.3          & 30.1          & \multicolumn{1}{c|}{23.5}          & 0.9          & 1.1          & \multicolumn{1}{c|}{1.0}          & 17.8          & 25.8          & \multicolumn{1}{c|}{21.1}          & 24.2          & 36.2          & 29.0          \\ \hline
\multicolumn{1}{c|}{Fusion}                   & 19.9          & \textbf{55.2}          & \multicolumn{1}{c|}{29.3}          & 28.6          & \textbf{73.3} & \multicolumn{1}{c|}{41.2}          & 31.2          & \textbf{78.6} & \multicolumn{1}{c|}{44.7}          & 1.1          & 3.8          & \multicolumn{1}{c|}{1.6}          & 16.9          & 50.0          & \multicolumn{1}{c|}{25.2}          & 24.4          & 72.5          & 36.5          \\ \hline
\multicolumn{19}{c}{Using predicted key clips}                                                                                                                                                                                                                                                                                                                                                                                                      \\ \hline
\multicolumn{1}{c|}{\citet{Kiddon2015MiseEP}} & 7.0           & 6.3           & \multicolumn{1}{c|}{6.6}           & 10.9          & 10.0          & \multicolumn{1}{c|}{10.4}          & 11.7          & 10.7          & \multicolumn{1}{c|}{11.2}          & 0.2          & 0.5          & \multicolumn{1}{c|}{0.3}          & 6.1           & 11.2          & \multicolumn{1}{c|}{7.9}           & 9.5           & 17.5          & 12.3          \\ \hline
\multicolumn{1}{c|}{SRL w/o heur.}            & 11.2          & 31.7          & \multicolumn{1}{c|}{16.6}          & 14.7          & 40.7          & \multicolumn{1}{c|}{21.6}          & 15.4          & 42.8 & \multicolumn{1}{c|}{22.6}          & 0.7          & \textbf{3.1} & \multicolumn{1}{c|}{1.2}          & 8.2           & \textbf{31.1} & \multicolumn{1}{c|}{13.0}          & 12.7          & \textbf{47.4} & 20.1          \\ \hline
\multicolumn{1}{c|}{SRL w/ heur.}             & \textbf{22.5}          & 29.9          & \multicolumn{1}{c|}{\textbf{25.7}}          & \textbf{26.2}          & 35.0          & \multicolumn{1}{c|}{\textbf{30.0}}          & \textbf{27.2}          & 36.3          & \multicolumn{1}{c|}{\textbf{31.1}}          & \textbf{0.9}          & 1.9          & \multicolumn{1}{c|}{\textbf{1.3}}          & \textbf{12.3}          & 23.1          & \multicolumn{1}{c|}{\textbf{16.1}}          & \textbf{18.8}          & 34.5          & \textbf{24.3}          \\ \hline
\multicolumn{1}{c|}{Visual}                   & 2.4           & 3.9           & \multicolumn{1}{c|}{3.0}           & 10.4          & 16.1          & \multicolumn{1}{c|}{12.6}          & 11.2          & 17.5          & \multicolumn{1}{c|}{13.7}          & 0.5          & 0.6          & \multicolumn{1}{c|}{0.6}          & 10.3          & 15.0          & \multicolumn{1}{c|}{12.2}          & 14.1          & 21.0          & 16.8          \\ \hline
\multicolumn{1}{c|}{Fusion}                   & 11.5          & \textbf{32.0}          & \multicolumn{1}{c|}{17.0}          & 16.6          & \textbf{42.5} & \multicolumn{1}{c|}{23.9}          & 18.1          & \textbf{45.6}          & \multicolumn{1}{c|}{25.9}          & 0.6          & 2.2          & \multicolumn{1}{c|}{1.0}          & 9.8           & 29.0          & \multicolumn{1}{c|}{14.6}          & 14.1          & 42.1          & 21.2          \\ \hline
\end{tabular}%
}
\vspace{-3mm}
\end{table*}

In this section, we evaluate results of the key clip selection described in \S\ref{sec:keyclipbaselines}. 
We evaluate using the accuracy, precision, recall and F1 score for the binary classification problem of whether a given clip in the video is a key clip. 
The results are shown in Table \ref{tab:keysent}.
We compare parsing-based heuristic models and supervised neural models, with ablations (model details in Appendix~\ref{app:neural_selection_model}).
From the experimental results in Table~\ref{tab:keysent}, we can see that: 
\begin{enumerate}[leftmargin=5mm]
\setlength\itemsep{-2mm}
    \item Unsupervised heuristic methods perform worse than neural models with training data. 
    This is despite the fact that the dataset used for training neural models has a different data distribution and domain from the test set.
    \item Among heuristic methods, pretrained SRL is better than \citet{Kiddon2015MiseEP} even though the second is trained on transcript text from YouCook2 videos. One possible reason is that the unsupervised segmentation method was specially designed for recipe texts, which are mostly simple, concise and imperative sentences found in recipe books, while the transcript is full of noise and tends to have longer, more complicated, and oral-style English.
    \item Post-processing significantly improves the SRL model, showing that filtering unrelated arguments and incorporating the cooking lexicon helps, especially with reducing false positives.
    \item Among neural method ablations, the model using only visual features performs worse than that using only text features. 
    The best model for identifying key clips among proposed baselines uses both visual and text information in the neural model.
\end{enumerate}

Besides quantitative evaluation, we analyzed key clip identification results and found a number of observations. 
First, background introductions, advertisements for the YouTube channel, etc.~can be relatively well classified due to major differences both visually and textually from procedural clips. 
Second, alignment and grounding between the visual and textual domains is crucial for key clip prediction, yet challenging. 
For example, the clip with the transcript sentence ``add more pepper according to your liking'' 
is identified as a key clip. However, it is in fact merely a suggestion made by the speaker about an imaginary scenario, rather than a real action performed and thus should not be regarded as a key procedure.

\subsection{Extraction Stage 2: Structured Procedure Extraction}
\label{sec:extractioneval}
In this stage, we perform key clip-level evaluation for structured procedural knowledge extraction by matching the ground truth and predicted structures with both exact match and two fuzzy scoring strategies.
To better show how stage 1 performance affects the whole pipeline, we evaluate on both ground truth (oracle) and predicted key clips. 
Similarly to the evaluation of key clip selection, we compare the parsing-based methods (\S\ref{sec:extract_utterances}), as well as purposing the action detection results from video signals for our task. 
Besides, we compare utterance-only and video-only baselines with our naive multi-modal fusion method.

We evaluate with respect to precision, recall and the F1 measure. 
Similarly to the evaluation method used for SRL~\cite{carreras2004introduction},
precision (P) is the proportion of verbs or arguments predicted by a
model which are correct, \ie $TP/\#\text{predicted}$ where $TP$ is the number of true positives. 
Recall (R) is the proportion of
correct verbs or arguments which are predicted by a model, \ie $TP/\#\text{gold}$. 
The key here is how to calculate $TP$ and we propose 3 methods: exact match, fuzzy matching, and partial fuzzy matching.
The first is straight forward, we count true positives if and only if the predicted phrase is an exact string match in the gold phrases.
However, because our task lies in the realm of open phrase extraction without predefined labels,
it is unfairly strict to count only the exact string matches as $TP$.
Also by design, the gold extraction results cannot always be found in the original transcript sentence (refer to \S\ref{sec:analysis}), 
so we are also unable to use token-based metrics as in sequence tagging~\cite{sang2003introduction}, or span-based metrics as in some question answering tasks~\cite{rajpurkar2016squad}.
Thus for the second metric we call \emph{``fuzzy''}, we leverage edit distance to enable fuzzy matching and assign a ``soft'' score for $TP$.
In some cases, the two strings of quite different lengths will hurt the \emph{fuzzy} score due to the nature of edit distance, even though one string is a substring of another.
To get around this, we propose a third metric, \emph{``partial fuzzy''} to get the score of the best matching substring with the length of the shorter string in comparison. 
Note that this third metric will bias towards shorter, correct phrases and thus we should have a holistic view of all 3 metrics during the evaluation.
Details of two fuzzy metrics are described in Appendix~\ref{app:fuzzy_matching}.
Table~\ref{tab:verbextract} illustrates evaluation results:
\begin{enumerate}[leftmargin=5mm]
\setlength\itemsep{-1mm}
\item Argument extraction is much more challenging compared to verb extraction, according the results: arguments contain more complex types of phrases (\eg objects, location, time, etc.) and are longer in length.
It is hard to identify complex arguments with our current heuristic or unsupervised baselines and thus the need for better supervised or semi-supervised models.
\item Heuristic SRL methods perform better than the unsupervised segmentation model even though the second is trained on our corpus.
This demonstrates the generality of SRL models, but the heuristics applied at the output of SRL models still improve the performance by reducing false positives.
\item The visual-only method performs the worst, mainly because of the domain gap between visual detection model outputs and our annotated verbs and arguments. 
Other reasons include: the closed label set predefined in EpicKitchen; challenges in domain transferring from closed to open extraction; different video data distribution between EpicKitchen (for training) and our dataset (YouCook2, for testing);  
limited performance of video detection model itself. 
\item Naive multimodal fusion leads to an overall performance drop to below the utterance-only model, 
partly due to the differences in video data distribution and domain, as well as the limitation of the predefined set of verbs and nouns in the EpicKitchen dataset, implying the need for better multimodal fusion method. 
Unsurprisingly, the recall for verb extraction raises after the fusion, suggesting that action detection in videos helps with the coverage. 
The drop in argument extraction suggests the complexity of arguments in our open extraction setting: it should be more than mere object detection.
\end{enumerate}

\noindent Besides quantitative results, we also showcase qualitative analysis of example extraction outputs in Appendix~\ref{app:casestudy}. 
From both, we suggest that there are two key challenges moving forward:\\
\textbf{Verb extraction:} We find that verb ellipsis is common in transcripts. 
The transcript text contains sentences where key action ``verbs'' do not have verb part-of-speech in the sentence.
For example, in the sentence ``give it a flip ...'' with the annotation (``flip'', ``pancake''), the model detects ``give'' as the verb rather than ``flip''.
Currently all our baselines are highly reliant on a curated lexicon for verb selection and thus such cases will get filtered out. 
How to deal with such cases with general verbs like \emph{make}, \emph{give}, \emph{do} remains challenging and requires extracting from the contexts. 

\noindent\textbf{Argument extraction:}
Speech-to-text errors are intrinsic in automatically acquired transcripts and cause problems during parsing that cascade.
Examples are that ``add flour'' being recognized as ``add flower'' and ``sriracha sauce'' being recognized as ``sarrah cha sauce'' causing wrong extraction outputs.
Coreference and ellipsis are also challenging and hurting current benchmark performance, as our baselines do not tackle any of these explicitly. 
Visual co-reference and language grounding~\cite{huang-buch-2018-finding-it,huang2017unsupervised} provides a feasible method for us to tackle these cases in the future.

\vspace{-2mm}
\section{Related Work}
\vspace{-2mm}
\label{sec:related}
\noindent \textbf{Text-based procedural knowledge extraction.}
Procedural text understanding and knowledge extraction~\cite{chu2017distilling,park2018learning,Kiddon2015MiseEP,jermsurawong2015predicting,liu2016jointly,long2016simpler,maeta2015framework,malmaud2014cooking,artzi2013weakly,kuehne2017weakly} has been studied for years on step-wise textual data such as WikiHow. 
\citet{chu2017distilling} extracted open-domain knowledge from how-to communities. Recently \citet{zhukov2019cross} also studied to adopt the well-written how-to data as weak supervision for instructional video understanding.
Unlike existing work on action graph/dependency extraction \cite{Kiddon2015MiseEP,jermsurawong2015predicting}, our approach differs as we extract knowledge from the visual signals and transcripts directly, not from imperative recipe texts.

\noindent \textbf{Instructional video understanding.} Beyond image semantics~\cite{yatskar2016}, unlike existing tasks for learning from instructional video~\cite{zhou2018end,tang2019coin,alayrac2016unsupervised,song2015tvsum,sener2015unsupervised,huang2016connectionist,sun2019videobert,1906.05743,plummer2017enhancing,palaskar2019multimodal}, combining video \& text information in procedures~\cite{yagcioglu2018recipeqa,2005.03684}, visual-linguistic reference resolution~\cite{huang-buch-2018-finding-it,huang2017unsupervised}, visual planning~\cite{Chang2019ProcedurePI}, joint learning of object and actions~\cite{zhukov2019cross,richard2018action,gao2017tall,damen2018scaling}, pretraining joint embedding of high level sentence with video clips~\cite{sun2019videobert,miech19howto100m}, our task proposal requires explicit structured knowledge tuple extraction. 

In addition to closely related work (\S \ref{sec:dataset}) there is a wide literature~\cite{malmaud2015s,zhou2018towards,ushiku2017procedural,nishimura2019frame,tang2019coin,huang2016connectionist,shi2019dense,ushiku2017procedural} that aims to predict/align dense procedural captions given the video, which are the most similar works to ours. 
\citet{zhou2018end} extracted temporal procedures and then generated captioning for each procedure. \citet{sanabria2018how2} proposes a multimodal abstractive summarization for how-to videos with either human labeled or speech-to-text transcript. \citet{alayrac2016unsupervised} also introduces an unsupervised step learning method from instructional videos. Inspired by cross-task sharing~\cite{zhukov2019cross}, which is a weakly supervised method to learn shared actions between tasks, fine grained action and entity are important for sharing similar knowledge between various tasks. We focus on \textit{structured} knowledge of fine-grained actions and entities.
Visual-linguistic coreference resolution~\cite{huang-buch-2018-finding-it,huang2017unsupervised} is among one of the open challenges for our proposed task. 

\section{Conclusions \& Open Challenges}
We propose a multimodal open procedural knowledge extraction task, present a new evaluation dataset, produce benchmarks with various methods, and analyze the difficulties in the task.
Meanwhile we investigate the limit of existing methods and many open challenges for procedural knowledge acquisition, including: to better deal with cases of coreference and ellipsis in visual-grounded languages;
exploit cross-modalities of information with more robust, semi/un-supervised models;
potential improvement from structured knowledge in downstream tasks (e.g., video captioning).
\bibliographystyle{acl_natbib}
\bibliography{references}

\clearpage

\appendix
\section{Comparison with Existing Datasets}
\label{app:existing_datasets}
There are publicly available datasets related to understanding instructional videos:
\begin{itemize}
    \item AllRecipes~\cite{Kiddon2015MiseEP} (AR). The authors collected 2,456 recipes from AllRecipes weibsite\footnote{\url{https://www.allrecipes.com/}}.
    The sentences in the dataset are mostly simple imperative English describing concise steps to make a given dish, where the first word is usually the verb describing the action. The ingredient list information is also available. In contrast, our task seeks to extract procedural information from more noisy, oral and erroneous languages in real life video context. 
    \item YouCook2\footnote{\url{http://youcook2.eecs.umich.edu/}}~\cite{zhou2018towards} (YC2). 
    The procedure steps for each video are annotated with temporal boundaries in the video and described by human-written imperative English sentences. However, this dataset does not contain more fine-grained annotations in a structured form. 
    \item HowTo100M\footnote{\url{https://www.di.ens.fr/willow/research/howto100m/}}\cite{miech19howto100m}. This is a large scale how-to videos dataset, searched on YouTube using the task taxonomy on WikiHow\footnote{\url{https://www.wikihow.com}} as a source. However, it does not contain any annotations although the domain is more general.
    \item CrossTask\footnote{\url{https://github.com/DmZhukov/CrossTask}}\cite{zhukov2019cross} (CT). 
    Based on HowTo100M, this dataset is used for weakly supervised learning with 18 tasks fully labeled and 65 related tasks unlabeled. Although the dataset is annotated in a structured way by separating verbs and objects, the label space is closed with predefined sets of verbs and objects. The dataset also does not allow multiple verbs or objects to be extracted for a single segment. 
    \item COIN\footnote{\url{https://coin-dataset.github.io/}}~\cite{tang2019coin}. This contains instructional (how-to) videos, in a closed taxonomy of tasks and steps. 
    The authors annotated time spans of steps in a video with pre-defined steps, however the biggest drawback is that it is unstructured and closed domain.
    \item How2\footnote{\url{https://github.com/srvk/how2-dataset}}~\cite{sanabria2018how2}. This dataset annotates ground truth transcript text to help abstractive summarization, a very different task than ours of structured data extraction. 
    \item HAKE\footnote{\url{http://hake-mvig.cn}}~\cite{li2019hake}.
    Human Activity Knowledge Engine (HAKE) is a large-scale knowledge base of human activities, built upon existing activity datasets, and supplies human instance action labels and corresponding body part level atomic action labels.
    However, HAKE uses closed activity and part state classes. It also does not contain videos of activities accompanied with narrative transcripts.
    \item TACOS\footnote{\url{http://www.coli.uni-saarland.de/projects/smile/page.php?id=tacos}}~\cite{regneri2013grounding}.
    This dataset considers the problem of grounding sentences describing actions in visual information extracted from videos in kitchen settings.
    The dataset contains expert annotations of low level activity tags, with a total of 60 different activity labels with numerous associated objects, and sequences of NL sentences describing actions in the kitchen videos.
    This dataset also does not support open extraction and the videos are provided using human annotated caption sentences, rather than transcript texts with noise.
\end{itemize}

\section{Neural Selection Model}
\label{app:neural_selection_model}
\begin{figure}[t]
    \centering
    \includegraphics[width=0.9\linewidth]{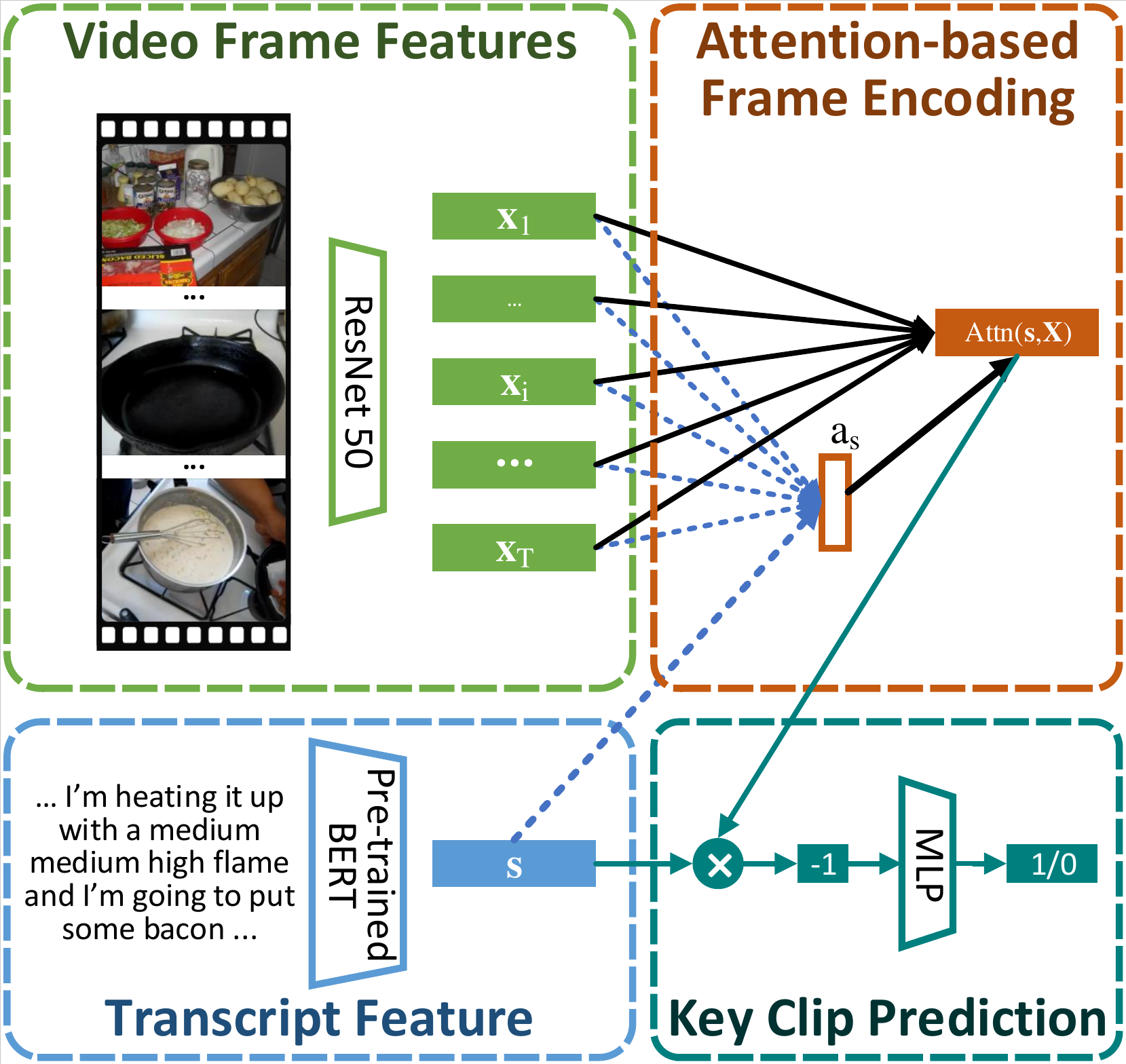}
    \caption{Neural key clip selection model.}
    \label{fig:sentence_selection_full_structure}
\end{figure}

Figure \ref{fig:sentence_selection_full_structure} presents the overall detailed structure of the neural selection model for combining utterance and video information for key clip selection.

\paragraph{Sentence token encoding}
Each input clip is accompanied with a sentence $S=\left\{t_1,\dots,t_k\right\}$ which has $k$ tokens. 
We use a pre-trained BERT~\cite{devlin2018bert} model as the encoder and extract the sentence representation $\mathbf{s}$.

\paragraph{Video frame features}
For each clip we uniformly sample $T=10$ frames and use an ImageNet-pretrained~\cite{deng2009imagenet} ResNet50~\cite{he2016deep} to extract the feature vector of each frame as $X=\left\{x_1,\cdots,x_T\right\}$.

\paragraph{Attention-based frame encoding}
To model the interaction between the encoded sentence and the feature of each frame, we adopt an attention-based method. We first calculate the attention weight $\mathbf{a}_s$ by a tensor product of sentence feature $\mathbf{s}$ with each video frame $\mathbf{x}_i$ followed by a softmax layer. 
Then we perform a weighted sum on all frame features to get $\text{Attn}(s,\mathbf{X})$.

\paragraph{Visual-utterance fusion}
Finally, we fuse the extracted transcript features $s$ with the attended video features $\text{Attn}(s,\mathbf{X})$ by a tensor product and flatten it into a vector. 
Then we use a non-linear activation layer to map these features into a real number, which represents the probability of the clip being a key clip.

\paragraph{Experiment details}
In the presented experiments, we use a pre-trained BERT~\cite{devlin2018bert} model\footnote{\url{https://github.com/hanxiao/bert-as-service}} to extract the continuous representation of each sentence.
During fine-tuning, the model is optimized by Adam optimizer~\cite{kingma2014adam} with the starting learning rate of $1e-4$.
The model is trained in a supervised fashion with a separate key clip/sentence classification dataset that is \emph{not} related to YouCook2. 
This auxiliary dataset will also be publicly released.
All of them are general domain instructional videos harvested from from YouTube. 
Human annotators labeled whether it is a key clip when given a video clip-sentence pair. 
In the end, we have 1,034 videos (40,146 pairs) for training the classification model. 
We split the dataset into two subset as 772 videos (28,519 pairs) and 312 videos (11,627 pairs) for training and validation (hyper-parameter tuning) respectively. 
The testing set is our proposed dataset with key clips and sentences annotated (see \S\ref{sec:dataset}), containing 356 videos and 15,523 pairs. 
The testing set used is the same as all other compared methods.

\section{SRL Argument Filtering}
\label{app:argument_filtering}
The argument types that we deem to not contribute as the procedural knowledge for completing the task and filter out include: 
ARG0 (usually refers to the subject, usually a person), AM-MOD (modal verb), AM-CAU (cause), AM-NEG (negation marker), AM-DIS (discourse marker), AM-REC (reciprocal), AM-PNC/PRP (purpose), AM-EXT (extent), and R-ARG* (in-sentence references).

\section{Fuzzy Matching and Partial Fuzzy Matching}
\label{app:fuzzy_matching}
\paragraph{Fuzzy matching}
Denote the Levenshtein distance between string $a$ and string $b$ as $d(a, b)$.
We then define a normalized pairwise score between 0 to 1 as $s(a, b) = d(a, b)/ max\{|a|, |b|\}$
Given a set of $n$ predicted phrases $X = \{x_1, ..., x_n\}$ and a set of $m$ ground truth phrases $G = \{g_1, ..., g_m\}$, we can find a set of $min(n, m)$ string pairs between predicted $X$ and ground truth $G$, as $M = \{(x_i, g_j)\}$ that maximizes the sum of scores $\sum_{(x_i, g_j) \in M} s(x_i, g_j)$. 
This assignment problem can be solved efficiently with Kuhn-Munkres~\cite{munkres1957algorithms} algorithm\footnote{\url{http://software.clapper.org/munkres/}}.
Since this fuzzy pairwise score is normalized, it can be regarded as a soft version for calculating $TP = max \sum_{(x_i, g_j) \in M} s(x_i, g_j)$.

\paragraph{Partial fuzzy matching}
The only difference from \emph{``fuzzy''} matching is that the scoring function now follows the  ``best partial'' heuristic that assuming the shorter string $a$ is length $|a|$, and the longer string $b$ is length $|b|$, we now calculate the score between shorter string and the best \emph{``fuzzy''} matching length-$|a|$ substring.
\begin{equation*}
\begin{aligned}
s(a,b) = max\{d(a, t)\}/ |a|, & t \in \text{substring of}~b,\\
& |t| = |a|, |a| < |b|
\end{aligned}
\end{equation*}
Both fuzzy metric implementations are based on FuzzyWuzzy\footnote{\url{https://github.com/seatgeek/fuzzywuzzy}}.

\section{Example Extractions}
\label{app:casestudy}
In this section, we showcase some example extractions from our overall best-performing baseline model ``SRL w/ heur.'' in Table~\ref{tab:casestudy} (next page).
We show both the annotated structured extractions as well as the model output, for a recipe of type \textit{``pizza marghetta''}.\footnote{\url{https://www.youtube.com/watch?v=FHvZgt3ExDI}}
Note that the table includes only transcript sentences that are annotated as key steps.
We can see that some sentences do no have extraction output from the model, while others tend to be over-extracted as long spans of text or incorrect due to the noisy nature of the transcript, degrading the extraction quality.
Verbs are also relatively better extracted than arguments for the proposed model.
From the extraction examples, we can also see that the model sometimes omits important action verbs in the extraction, and extract pronouns like ``it'', ``this'', ``here'', etc. as arguments. 
This suggests that the utterance-only model cannot handle \emph{coreference} and \emph{ellipsis} scenarios very well, which is one of the key difficulties of the proposed task (discussed in Section~\ref{sec:analysis}).
This also partly implies the need for utilizing the visual information to extract actions and arguments that are not included in transcript, as well as visual co-reference and language grounding to help resolve what objects the pronouns in the extracted arguments are referring to.

\section{Reproducibility Details}
\label{app:repro}
All the experiments in this paper are performed on a workstation with one 8-core Intel i7 processor, 32GB RAM and one NVIDIA Tesla K80 GPU.

The SRL-based models' runtime are bounded by the inference speed of the pretrained model, and take 30 minutes on average to complete the evaluation.
The code is based on the semantic role labeling pretrained model from \url{https://github.com/allenai/allennlp-hub/blob/master/allennlp_hub/pretrained/allennlp_pretrained.py}.

\citet{Kiddon2015MiseEP} models are trained from scratch with default hyperparameter settings on our dataset using only CPU, and take 2 hours on average to complete training and evaluation.
We directly use the released code (\url{https://github.com/uwnlp/recipe-interpretation}) to train the unsupervised model on our dataset for evaluation.

The neural selection model (Section~\ref{sec:neuralsel}) takes 6 hours on average to train on both transcript and video data.
The number of parameters for the neural model is the addition of those from BERT-base model, ResNet-50 model, an attention module and a MLP layer.
More details are described in Appendix~\ref{app:neural_selection_model}.

The action and object detection model (Section~\ref{sec:videodetection}) pretrained on EpicKitchen dataset is retrieved from \url{https://github.com/epic-kitchens/action-models}.
In our use case, we only need to perform the inference steps on our dataset, which costs 3 hours on average for all the video clips in our proposed dataset.

\clearpage
\onecolumn

\small
\begin{longtable}{>{\raggedright}p{5cm}|>{\raggedright}p{2cm}|>{\raggedright}p{1cm}|>{\raggedright}p{2.5cm}|>{\raggedright}p{1cm}|>{\raggedright\arraybackslash}p{2.5cm}}
\caption{
Example extractions compared with gold annotations for a recipe of type ``pizza marghetta''. Only transcript sentences that are annotated as key steps are included. Some long and incorrect extractions are quoted with omission.}
\label{tab:casestudy}
\\
Transcript Sentence                                                                                                                                                                                                                                                                                                                                                                                                                                                                                                                                      & Procedure Summary                                                  & Gold Verbs       & Gold Arguments                                       & Extracted Verbs           & Extracted Arguments                                                                                                                                                                 \\ \hline
so we 've placed the dough directly into   the caputo flour that we import from italy.                                                                                                                                                                                                                                                                                                                                                                                                                                                                   & place dough in caputo flour                                        & place            & dough, caputo flour                                  & place                     & so, the dough, directly, into the caputo flour that we import from italy                                                                                                            \\ \hline
and then we give it a flip as i 've read   in some, some manuals for italian pizza that are neapolitan style.                                                                                                                                                                                                                                                                                                                                                                                                                                            & flip dough                                                         & flip             & dough                                                &                           &                                                                                                                                                                                     \\ \hline
but that 's what we do anyway, we   sprinkle the surface and then real quick.                                                                                                                                                                                                                                                                                                                                                                                                                                                                            & sprinkle the surface                                               & sprinkle         & surface                                              & sprinkle                  & the surface                                                                                                                                                                         \\ \hline
we just give a squish with our palm and   make it flat in the center.                                                                                                                                                                                                                                                                                                                                                                                                                                                                                    & squish dough with palm; flatten center                             & squish; flatten  & dough, with palm; center of dough                    & make                      & just, it flat in the center                                                                                                                                                         \\ \hline
we dimple the rest of the pizza, moving   the pizza around definitely handmade, definitely handmade, not trying your   best not to disturb the edge 'cause that 's where you 're going to get a lot   of natural bubbles from the fermentation.                                                                                                                                                                                                                                                                                                          & dimple the rest of pizza; move the pizza around                    & dimple; move     & pizza; pizza                                         & move                      & the pizza, around, ``definitely handmade, definitely handmade''                                                                                                                      \\ \hline
once you 've dimpled, you 're going to do   a quick stretch, while you 're rotating.                                                                                                                                                                                                                                                                                                                                                                                                                                                                     & stretch dough;rotate dough                                         & stretch; rotate  & dough; dough                                         & rotate                    & you                                                                                                                                                                                 \\ \hline
and then we 're going to put it, pick it   up and put it on the backs of our hands and just let it kind of hang down on   the backs of our knuckles.                                                                                                                                                                                                                                                                                                                                                                                                     & pick up dough and let it hang on backs of knuckles                 & pick up; hang on & dough; dough, backs of knuckles                      & put; pick; put;  let      & it; it; it, on the backs of our hands; just, it kind of hang down on the   backs of our knuckles                                                                                    \\ \hline
do you want a quarter to two and a half   ounces of sauce, and i 'm going to put the cheese on.                                                                                                                                                                                                                                                                                                                                                                                                                                                          & put cheese on                                                      & put on           & cheese                                               & put                       & the cheese, on                                                                                                                                                                      \\ \hline
olive oil actors come out of the oven,   and we put on here.                                                                                                                                                                                                                                                                                                                                                                                                                                                                                             & put on fresh basil                                                 & put on           & fresh basil                                          & put                       & here                                                                                                                                                                                \\ \hline
so that 's all we do to the pizza and   then we go into the oven just going to place it directly on the stone and   give it a shake.                                                                                                                                                                                                                                                                                                                                                                                                                     & place pizza in oven, give it a shake                               & place; shake     & pizza, oven; pizza                                   & place                     & it, directly on the stone                                                                                                                                                           \\ \hline
so will have to rotate it every thirty to   forty five seconds, your home oven will take you about ten minutes.                                                                                                                                                                                                                                                                                                                                                                                                                                          & rotate pizza every 30-45 seconds                                   & rotate           & pizza, 30-45 seconds                                 & rotate, take              & it, every thirty to forty five seconds; will, you, about ten minutes                                                                                                                \\ \hline
we 're going to put a little bit of extra   virgin olive oil directly on.                                                                                                                                                                                                                                                                                                                                                                                                                                                                                & put extra vigrin olive oil on                                      & put on           & extra virgin olive oil                               & put                       & a little bit of extra virgin olive oil, directly, on                                                                                                                                \\ \hline
yeah , over the whole pizza and then in   america we cut it, but i 'm told in, uh, in italy you cut it yourself.                                                                                                                                                                                                                                                                                                                                                                                                                                         & cut the pizza                                                      & cut              & pizza                                                & cut, cut                  & in america, it, in; in italy, it, yourself                                                                                                                                          \\ \hline
and we start to knead, so we need about   though for this fifteen twenty minutes until we have a very nice texture.                                                                                                                                                                                                                                                                                                                                                                                                                                      & knead dough for 15-20 minutes                                      & knead            & dough, 15-20 minutes                                 &                           &                                                                                                                                                                                     \\ \hline
you make a nice bowl with cover, the door   to avoid crusting an you let it rise at room temperature for three four hour.                                                                                                                                                                                                                                                                                                                                                                                                                                & make bowl with cover, rise dough at room temperature for 3-4 hours & make; rise       & bowl, with cover; dough, room temperature, 3-4 hours & make; let                 & a nice bowl, with cover; it rise at room temperature for three four hour                                                                                                            \\ \hline
one sourdough, as almost double is volume   we make six small balls.                                                                                                                                                                                                                                                                                                                                                                                                                                                                                     & make six small balls                                               & make             & six small balls                                      & make                      & one sourdough, as almost double is volume, six small balls                                                                                                                          \\ \hline
so we make nice round shape now.                                                                                                                                                                                                                                                                                                                                                                                                                                                                                                                         & make a round shape                                                 & make             & round shape                                          & make                      & so, nice round shape, now                                                                                                                                                           \\ \hline
we cover again with film to avoid the   door crafting an we elect, leave at room times or for a couple of hour or if   you like you can place in a warm place for forty five minutes around pizza   dough is ready.                                                                                                                                                                                                                                                                                                                                      & cover with film                                                    & cover            & film                                                 & cover; leave; place       & again, with film, ``to avoid the door ... pizza dough is ready'';   at room times or for a couple of hour; in a warm place, for forty five   minutes, around                          \\ \hline
and we start making flat with our hands.                                                                                                                                                                                                                                                                                                                                                                                                                                                                                                                 & make the dough flat with hands                                     & make flat        & dough, with hands                                    & make                      & flat, with our hands                                                                                                                                                                \\ \hline
now we had a tomato and we spread all   over our pizza dough.                                                                                                                                                                                                                                                                                                                                                                                                                                                                                            & spread tomato over pizza dough                                     & spread           & tomato, pizza dough                                  & spread                    & all, over our pizza dough                                                                                                                                                           \\ \hline
we had the mozzarella cheese shredded   little basil pizza is ready to be back.                                                                                                                                                                                                                                                                                                                                                                                                                                                                          & add mozzarella cheese, shredded basil                              & add              & mozzarella cheese, shredded basil                    & shred                     & pizza                                                                                                                                                                               \\ \hline
now we should make pizza at around seven   hundred fifty degrees.                                                                                                                                                                                                                                                                                                                                                                                                                                                                                        & make pizza                                                         & make             & pizza, 750 degrees                                   & make                      & now, should, pizza, at around seven hundred fifty degrees                                                                                                                           \\ \hline
fahrenheit four five, seven finish our   pizza with basil and fresh olive oil.                                                                                                                                                                                                                                                                                                                                                                                                                                                                           & finish pizza with basil and olive oil                              & finish           & pizza, with basil, olive oil                         &                           &                                                                                                                                                                                     \\ \hline
OK , well, to make the margarita pizza we   're going to start off with by stretching the dough and we do make gardell   everyday fresh in our kitchen after that we 're going to add some shredded   mozzarella cheese.                                                                                                                                                                                                                                                                                                                                 & stretch the dough, add shredded mozzarella cheese                  & stretch;  add    & dough; shredded mozzarella cheese                    & make; stretch;  make; add & ``the margarita pizza we 're going to start off with'', by   stretching the dough; the dough; gardell, everyday, fresh, in our kitchen,   after that; some shredded mozzarella cheese \\ \hline
then our version of a margarita pizza has   four dollops of tomato sauce, along with some fresh, chopped tomato, then it   goes into the wood burning oven, they certainly, we 've been cooking with for   fifteen years.                                                                                                                                                                                                                                                                                                                                & put chopped tomato and tomato sauce into oven                      & put              & chopped tomato and tomato sauce, into oven           & oven; cook                & wood burning; with, for fifteen years                                                                                                                                               \\ \hline
after the margarita pizza comes out of   the oven, we finish it with some fresh, grated parmesan cheese.                                                                                                                                                                                                                                                                                                                                                                                                                                                 & finish with more cheese, fresh basil                               & finish           & pizza, with more cheese, fresh basil                 &                           &                                                                                                                                                                                     \\ \hline
here you can find how i do it out on the   website, and then i 've also got my basic classic tomato sauce plus that i   use for just about all my pizzas, an i 'm going to go ahead and lay that down   in a, in a nice coat and i like you know, probably frankly i probably like a   medium amount of sauce people, people have said little bit less a little bit   more i kind of like it a little bit right in the middle use, the back of the   spoon there, yeah, you spoon it out with the front, then just kind of use the   back spread it out. & lay down tomato sauce                                              & lay down         & tomato sauce                                         & lay; spoon;  spread       & that, down, ``in a , in a nice coat''; it, with the front; just,   kind of, it                                                                                                        \\ \hline
and then i like to use our leave about   half an inch all the way around the pizza.                                                                                                                                                                                                                                                                                                                                                                                                                                                                      & leave half an inch around pizza                                    & leave            & half an inch leaf, around pizza                      &                           &                                                                                                                                                                                     \\ \hline
i 'm going to go ahead and lay this down.                                                                                                                                                                                                                                                                                                                                                                                                                                                                                                                & lay down cheese                                                    & lay down         & cheese                                               & lay                       & this, down                                                                                                                                                                          \\ \hline
and then you can lay some basil down now.                                                                                                                                                                                                                                                                                                                                                                                                                                                                                                                & lay basil down                                                     & lay down         & basil                                                & lay                       & and, then, can, some basil, down, now                                                                                                                                               \\ \hline
i like to press down a little bit to make   sure that it stays down, stays down as it cooks if it stays down it.                                                                                                                                                                                                                                                                                                                                                                                                                                         & press basil down                                                   & press down       & basil                                                & press; make               & down, a little bit, ``to make sure ... it stays down it'';   ``sure ... it stays down it''                                                                                               \\ \hline
let 's go ahead and pop this in the oven,   so in the interest of full disclosure.                                                                                                                                                                                                                                                                                                                                                                                                                                                                       & pop pizza in oven                                                  & pop              & pizza, oven                                          & let; pop                  & ``'s go ahead and ... full disclosure''; this, in the oven, so in   the interest of full disclosure                                                                                   \\ \hline
i actually kept it on the pizza stone.                                                                                                                                                                                                                                                                                                                                                                                                                                                                                                                   & keep it on pizza stone                                             & keep             & pizza, pizza stone                                   & keep                      & actually, it, on the pizza stone                                                                                                                                                    \\ \hline
\end{longtable}%

\end{document}